%% file: main.tex
\documentclass[10pt,twocolumn,letterpaper]{article}

\usepackage{cvpr}              


\definecolor{cvprblue}{rgb}{0.21,0.49,0.74}
\usepackage[pagebackref,breaklinks,colorlinks,allcolors=cvprblue]{hyperref}

\usepackage{algorithm}   
\usepackage{algorithmic}
\usepackage{epsfig}
\usepackage{graphicx}
\usepackage{amsmath,bm}
\usepackage{amssymb}
\usepackage{bbding}
\usepackage{color}

\usepackage{booktabs}
\usepackage{enumitem}
\usepackage{cuted}
\usepackage{multirow}
\usepackage{xcolor}

\usepackage{bbding}
\usepackage[T1]{fontenc}

\def\paperID{*****} 
\def\confName{CVPR}
\def\confYear{2025}

\title{Direct3D-S2: Gigascale 3D Generation Made Easy with Spatial Sparse Attention}

\vspace{-16pt}
\author{%
Shuang Wu$^{1,2}$\thanks{Equal contribution. Work done during internship at DreamTech.} \quad 
Youtian Lin$^{1,2*}$ \quad
Feihu Zhang$^{2}$ \quad
Yifei Zeng$^{1,2}$ \quad
Yikang Yang$^{1}$ \quad
Yajie Bao$^{2}$ \quad \\
Jiachen Qian$^{2}$ \quad
Siyu Zhu$^{3}$ \quad   
Xun Cao$^{1}$ \quad 
Philip Torr$^{4}$ \thanks{Chief scientific advisor of DreamTech.} \quad
Yao Yao$^{1}$\thanks{Corresponding author.} 
\\ \vspace{-10pt}\\
$^{1}$Nanjing University \qquad 
$^{2}$DreamTech \qquad 
$^{3}$Fudan University \qquad
$^{4}$University of Oxford\\
}

\begin{document}
\maketitle
\begin{strip}
	\vspace{-60pt}
	\centering
	\includegraphics[width=1\textwidth]{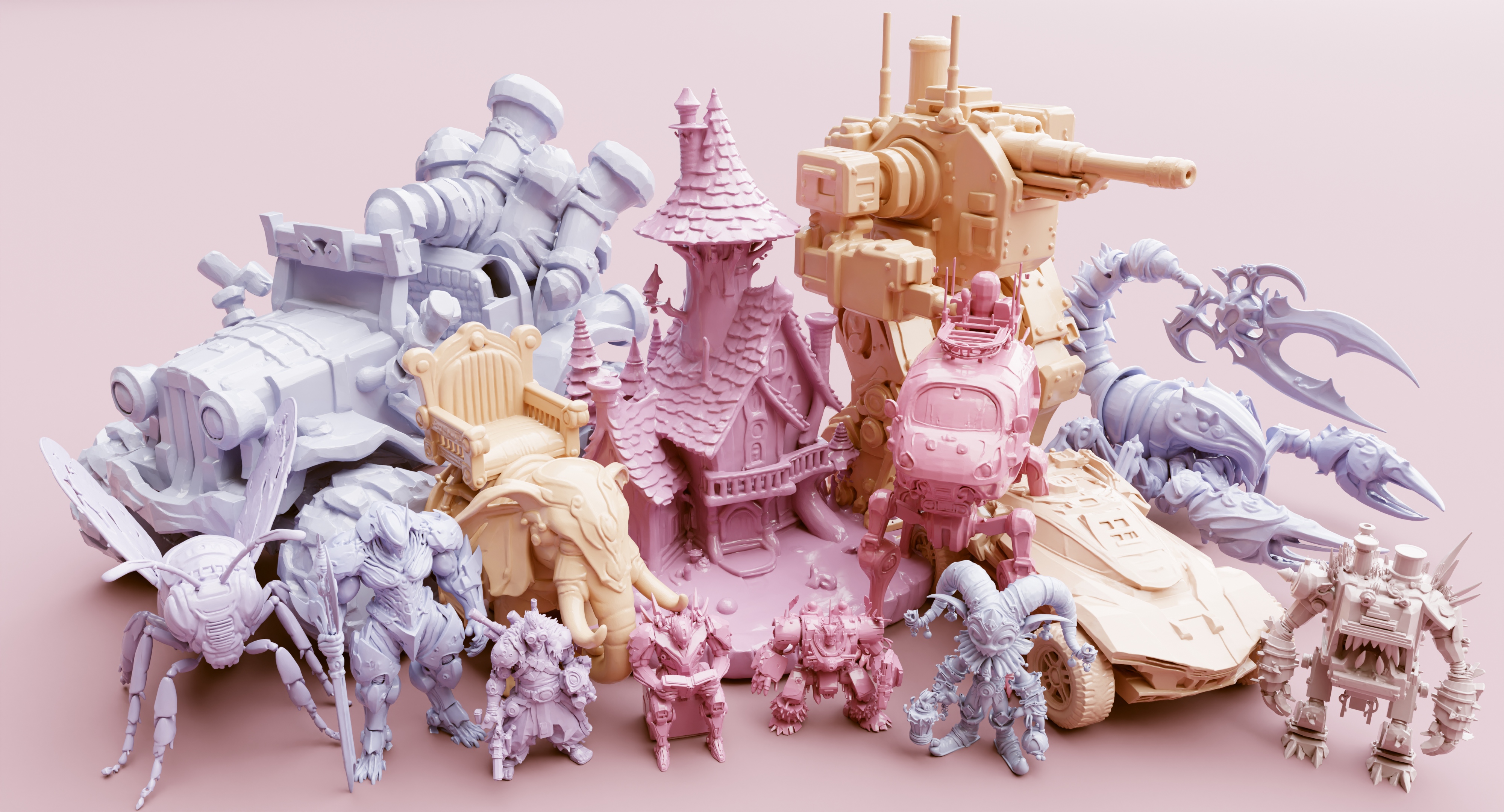}
    \vspace{-8pt}
    \captionsetup{type=figure,font=small,position=top}
    \caption{
        Mesh generation results from our method on different input images. Our method can generate detailed and complex 3D shapes. The meshes show fine geometry and high visual quality, demonstrating the strength of our approach for high-resolution 3D generation.
    } 
    \label{fig:teaser}
    \vspace{-14pt}
\end{strip}

\input{contents/0_abstract}

\input{contents/1_intro}

\input{contents/2_related_work}

\input{contents/3_method}

\input{contents/4_experiments}

\input{contents/5_conclusion}

{
    \small
    \bibliographystyle{ieeenat_fullname}
    \bibliography{main}
}

\begin{figure*}
  \centering
  \includegraphics[width=0.97\linewidth]{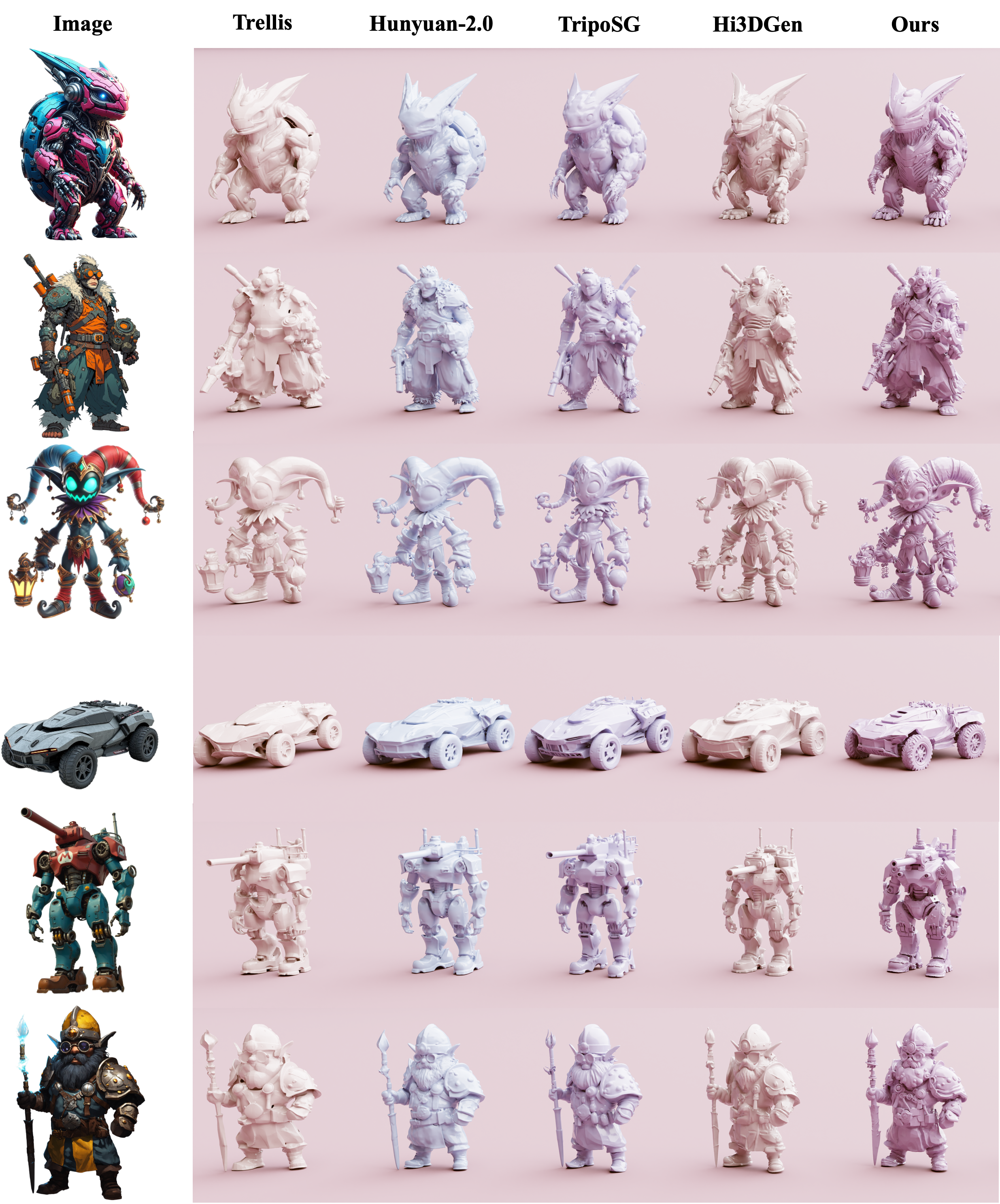}
  \caption{More qualitative comparisons between other \textbf{open-source} image-to-3D methods and our approach. \textbf{\emph{Best viewed with zoom-in.}}}
  \label{fig:mesh_compare}
\end{figure*}

\begin{figure*}
  \centering
  \includegraphics[width=0.97\linewidth]{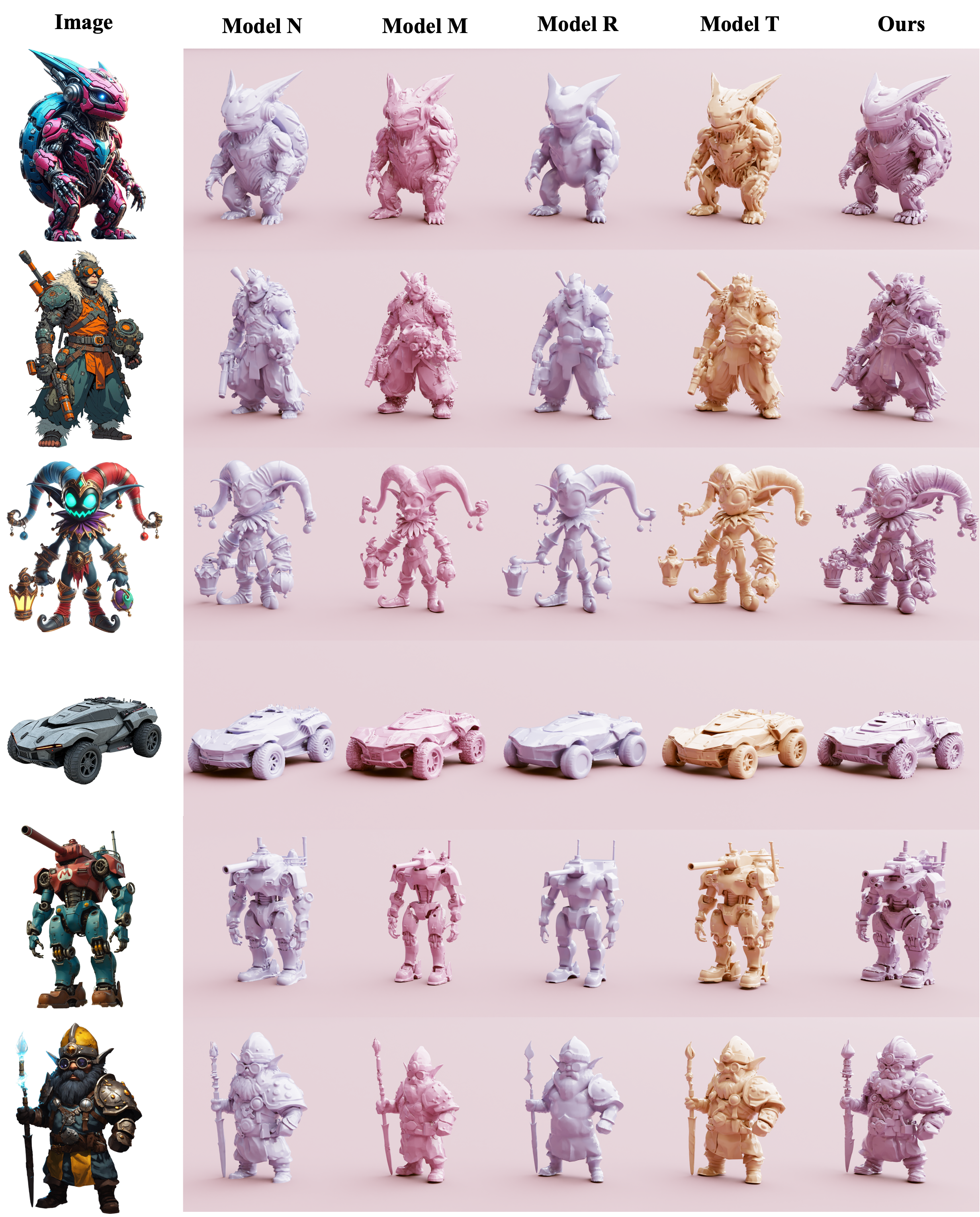}
  \caption{Qualitative comparisons between \textbf{closed-source commercial} image-to-3D models and our approach. Note that for each closed-source model we use the default setting of their web app. \textbf{\emph{Best viewed with zoom-in.}}}
  \label{fig:mesh_compare_c}
\end{figure*}

\vspace{-10pt}
\end{document}

%% file: contents/0_abstract.tex
\begin{abstract}
Generating high-resolution 3D shapes using volumetric representations such as Signed Distance Functions (SDFs) presents substantial computational and memory challenges. We introduce Direct3D-S2, a scalable 3D generation framework based on sparse volumes that achieves superior output quality with dramatically reduced training costs.
Our key innovation is the \textbf{Spatial Sparse Attention (SSA)} mechanism, which greatly enhances the efficiency of Diffusion Transformer (DiT) computations on sparse volumetric data. SSA allows the model to effectively process large token sets within sparse volumes, substantially reducing computational overhead and achieving a 3.9$\times$ speedup in the forward pass and a 9.6$\times$ speedup in the backward pass.
Our framework also includes a variational autoencoder (VAE) that maintains a consistent sparse volumetric format across input, latent, and output stages. Compared to previous methods with heterogeneous representations in 3D VAE, this unified design significantly improves training efficiency and stability.
Our model is trained on public available datasets, and experiments demonstrate that Direct3D-S2 not only surpasses state-of-the-art methods in generation quality and efficiency, but also enables \textbf{training at 1024³ resolution using only 8 GPUs}, a task typically requiring at least 32 GPUs for volumetric representations at $256^3$ resolution, thus making gigascale 3D generation both practical and accessible. Project page: \href{https://www.neural4d.com/research/direct3d-s2}{https://www.neural4d.com/research/direct3d-s2}.
\end{abstract}

%% file: contents/1_intro.tex
\section{Introduction}
\label{sec:intro}

Generating high-quality 3D models directly from text or images offers significant creative potential, enabling rapid 3D content creation for virtual worlds, product prototyping, and various real-world applications. This capability has garnered increasing attention across domains such as gaming, virtual reality, robotics, and computer-aided design.

Recently, large-scale 3D generative models based on implicit latent representations have made notable progress. 
These methods leverage neural fields for shape representation, benefiting from compact latent codes and scalable generation capabilities.
For instance, 3DShape2Vecset~\cite{zhang20233dshape2vecset} pioneered diffusion-based shape synthesis by using a Variational Autoencoder (VAE)~\cite{kingma2013vae} to encode 3D shapes into a latent vecset, which can be decoded into neural SDFs or occupancy fields and rendered via Marching Cubes~\cite{lorensen1998marching}. The latent vecset is then modeled with a diffusion process to generate diverse 3D shapes.
CLAY~\cite{zhang2024clay} extended this pipeline with Diffusion Transformers (DiT)~\cite{peebles2023scalable}, while TripoSG~\cite{li2025triposg} further improved fidelity through rectified flow transformers and hybrid supervision.
However, implicit latent-based methods often rely on VAEs with asymmetric 3D representations, resulting in lower training efficiency that typically requires hundreds of GPUs.

Explicit latent methods have emerged as a compelling alternative to implicit ones, offering better interpretability, simpler training, and direct editing capabilities, while also adopting scalable architectures such as DiT~\cite{peebles2023scalable}. For instance, Direct3D~\cite{wu2024direct3d} proposes to use tri-plane latent representations to accelerate training and convergence. XCube~\cite{ren2024xcube} introduces hierarchical sparse voxel latent diffusion for $1024^3$ sparse volume generation, but only restricted to millions of valid voxels, limiting the final output quality. Trellis~\cite{xiang2024structured} integrates sparse voxel representations of $256^3$ resolution, with the rendering supervision for the VAE training. In general, due to high memory demands, existing explicit latent methods are limited in output resolution. Scaling to $1024^3$ with sufficient latent tokens and valid voxels remains challenging, as the quadratic cost of full attention in DiT renders high-resolution training computationally prohibitive.

To address the challenge of high-resolution 3D shape generation, we propose Direct3D-S2, a unified generative framework that utilizes sparse volumetric representations. At the core of our approach is a novel Spatial Sparse Attention (SSA) mechanism, which substantially improves the scalability of diffusion transformers in high-resolution 3D shape generation by selectively attending to spatially important tokens via learnable compression and selection modules. Specifically, we draw inspiration from the key principles of Native Sparse Attention (NSA)~\cite{yuan2025native}, which integrates compression, selection, and windowing to identify relevant tokens based on global-local interactions. While NSA is designed for structurally organized 1D sequences, it is not directly applicable to unstructured, sparse 3D data. To adapt it, we redesign the block partitioning to preserve 3D spatial coherence and revise the core modules to accommodate the irregular nature of sparse volumetric tokens. This enables efficient processing of large token sets within sparse volumes. We implement a custom Triton~\cite{tillet2019triton} GPU kernel for SSA, achieving a \textbf{3.9$\times$ speedup in the forward pass} and a \textbf{9.6$\times$ speedup in the backward pass} compared to FlashAttention-2 at $1024^3$ resolution.

Our framework also includes a VAE that maintains a consistent sparse volumetric format across input, latent, and output stages. This unified design eliminates the need for cross-modality translation, commonly seen in previous methods using mismatched representations such as point cloud input, 1D vector latent, and dense volume output, thereby improving training efficiency, stability, and geometric fidelity. After the VAE training, the DiT with the proposed SSA will be trained on the converted latents, enabling scalable and efficient high-resolution 3D shape generation.

Extensive experiments demonstrate that our approach successfully achieves high-quality and efficient gigascale 3D generation, a milestone previously unattainable by explicit 3D latent diffusion methods. Compared to prior native 3D diffusion techniques, our model consistently generates highly detailed 3D shapes while considerably reducing computational costs. Notably, Direct3D-S2 requires only 8 GPUs to train on public datasets~\cite{deitke2023objaversexl, deitke2023objaverse, liu2023openshape} at a resolution of $1024^3$, in stark contrast to prior state-of-the-art methods, which typically require 32 or more GPUs even for training at $256^3$ resolution.

%% file: contents/2_related_work.tex
\section{Related work}

\subsection{Multi-view Generation and 3D Reconstruction}

Large-scale 3D generation has been advanced by methods such as~\cite{li2023instant3d, liu2023one2345, long2024wonder3d, xu2024instantmesh}, which employ multi-view diffusion models~\cite{wang2023imagedream} trained on 2D image prior models like Stable Diffusion~\cite{rombach2022high} to generate multi-view images of 3D shapes. These multi-view images are then used to reconstruct 3D shapes via generalized sparse-view reconstruction models. Follow-up works~\cite{liu2024one, lu2024direct2, tang2024lgm, xu2024grm, zhang2024gs} further improve the quality and efficiency of reconstruction by incorporating different 3D representations.
Despite these advances, these methods still face challenges in maintaining multi-view consistency and shape quality. The synthesized images may fail to faithfully represent the underlying 3D structure, which could result in artifacts and reconstruction errors.
Another limitation is the reliance on rendering-based supervision, such as Neural Radiance Fields (NeRF)~\cite{mildenhall2020nerf} or DMTet~\cite{NEURIPS2021_30a237d1}. While this avoids the need for direct 3D supervision (e.g., meshes), it adds significant complexity and computational overhead to the training process. Rendering-based supervision can be slow and costly, especially when scaled to large datasets.

\subsection{Large Scale 3D Latent Diffusion Model}

Motivated by recent advances in Latent Diffusion Models (LDMs)~\cite{rombach2022high} in 2D image generation, several methods have extended LDMs to 3D shape generation. These approaches broadly fall into two categories: vecset-based methods and voxel-based methods. Implicit vecset-based methods, such as 3DShape2Vecset~\cite{zhang20233dshape2vecset}, Michelangelo~\cite{zhao2024michelangelo}, CLAY~\cite{zhang2024clay}, and CraftsMan3D~\cite{li2024craftsman}, represent 3D shapes using latent vecset and reconstruct meshes through neural SDFs or occupancy fields. However, implicit methods are typically constrained by the size of vecset: larger vecset leads to more complex mappings to the 3D shape and requires longer training times. In contrast, voxel-based methods, such as XCube~\cite{ren2024xcube}, Trellis~\cite{xiang2024structured}, and more recent works~\cite{he2025triposf, ye2025hi3dgen}, employ voxel grids as latent representations, providing more interpretability and easier training. Nevertheless, voxel-based methods face limitations in latent resolution due to cubic growth in GPU memory requirements and high computational costs associated with attention mechanisms. To address this issue, our work specifically targets reducing the computational overhead of attention mechanisms, thereby enabling the generation of high-resolution voxel-based latent representations that were previously infeasible.

\subsection{Efficient Large Tokens Generation}

Generating large tokens efficiently is a challenging problem, especially for high-resolution data. Native Sparse Attention (NSA)~\cite{yuan2025native} addresses this by introducing adaptive token compression that reduce the number of tokens involved in attention computation, while maintaining performance comparable to full attention. NSA has been successfully applied to large language models~\cite{yuan2025native, pikekos2025mixture} and video generation~\cite{tan2025dsv}, showing significant reductions in attention cost. In this paper, we extend token compression to 3D data and propose a new Spatial Sparse Attention (SSA) mechanism. SSA adapts the core ideas of NSA but modifies the block partitioning strategy to respect 3D spatial coherence. We also redesign the compression, selection, and window modules to better fit the properties of sparse 3D token sets.
Another line of work, such as linear attention~\cite{katharopoulos2020transformers}, reduces attention complexity by approximating attention weights with linear functions. Variants of this technique have been applied in image~\cite{xie2024sana, zhu2024dig} and video generation~\cite{lu2022linear} to improve efficiency. However, the absence of non-linear similarity can lead to a significant decline in the performance of the model.

%% file: contents/3_method.tex
\section{Sparse SDF VAE}
\label{sec:SDF-VAE}

\begin{figure*}[!t]
  \centering
  \includegraphics[width=1\linewidth]{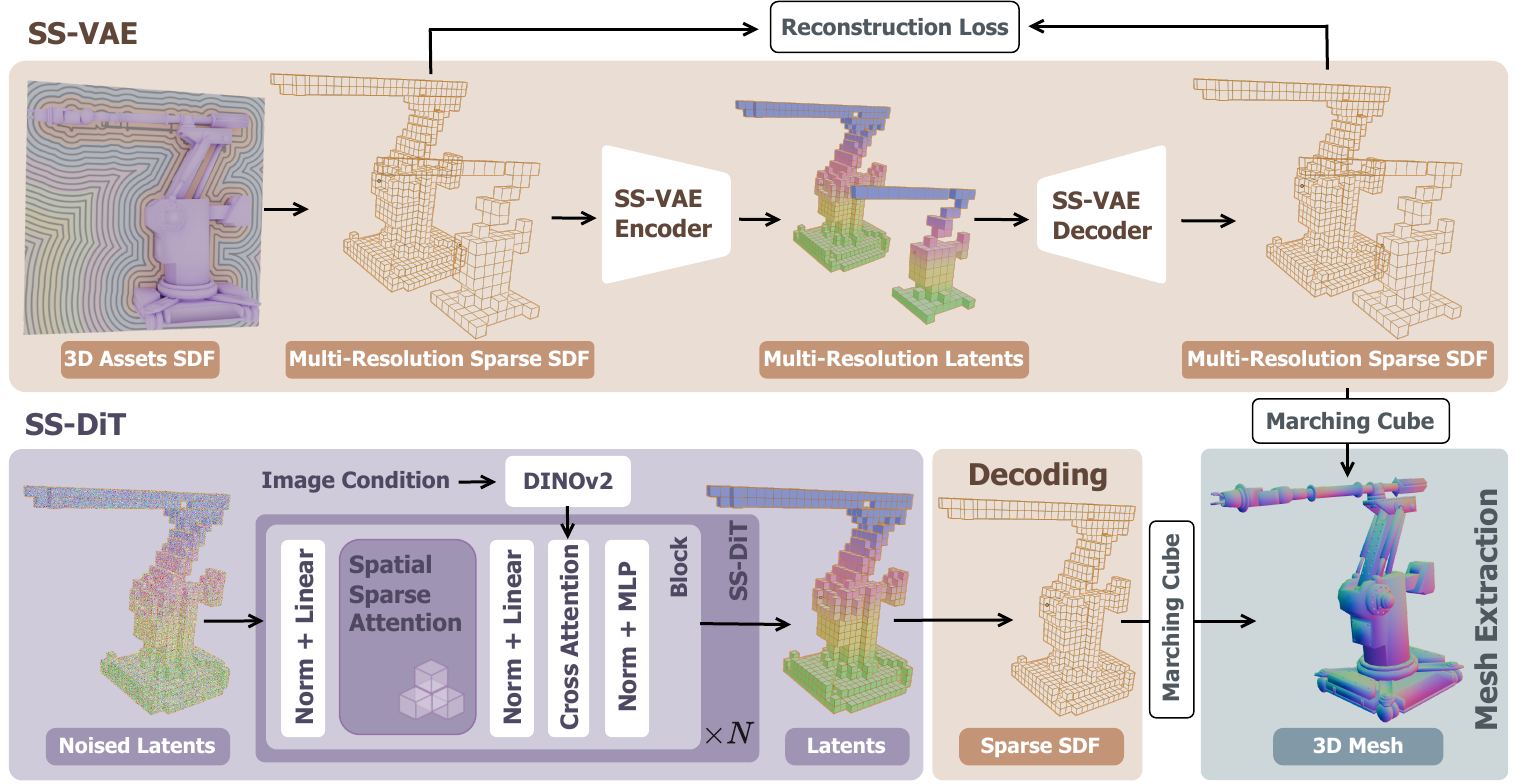}
  \caption{The framework of our Direct3D-S2. We propose a fully end-to-end sparse SDF VAE (SS-VAE), which employs a symmetric encoder-decoder network to efficiently encode high-resolution sparse SDF volumes into sparse latent representations $\mathbf{z}$. Then we train an image-conditioned diffusion transformer (SS-DiT) based on $\mathbf{z}$, and design a novel Spatial Sparse Attention (SSA) mechanism that substantially improves the training and inference efficiency of the DiT.}
  \label{fig:pipeline}
  \vspace{-8pt}
\end{figure*}

While variational autoencoders (VAEs) have become the cornerstone of 2D image generation by compressing pixel representations into compact latent spaces for efficient diffusion training, their extension to 3D geometry faces fundamental challenges. Unlike images with standardized pixel grids, 3D representations lack a unified structure, such as meshes, point clouds, and implicit fields, each requires specialized processing. This fragmentation forces existing 3D VAEs into asymmetric architectures with compromised efficiency. 
For instance, prominent approaches~\cite{chen2024dora, zhang2024clay,zhao2025hunyuan3d} based on vecset~\cite{zhang20233dshape2vecset} encode the input point cloud into a vector set latent space before decoding it into SDF field, while Trellis~\cite{xiang2024structured} and XCube~\cite{ren2024xcube} rely on differentiable rendering or neural kernel surface reconstruction~\cite{huang2023neural} to bridge their latent spaces to usable meshes. 
These hybrid pipelines introduce computational bottlenecks and geometric approximations that limit their scalability to high-resolution 3D generation. 
In this paper, we propose a fully end-to-end sparse SDF VAE that employs a symmetric encoding-decoding network to encode high-resolution sparse SDF volumes into a sparse latent representation, substantially improving training efficiency while maintaining geometric precision.

Given a mesh represented as a signed distance field (SDF) volume V with resolution $R^3$ (e.g., $ 1024^3 $), the SS-VAE first encodes it into a latent representation $ \mathbf{z} = E(V) $, then reconstructs the SDF through the decoder $ \tilde{V} = D(\mathbf{z})$. Direct processing of dense $ R^3 $ SDF volumes proves computationally prohibitive. To address this, we strategically focus on valid sparse voxels where absolute SDF values fall below threshold $\tau$: 
\begin{equation}
    V = \{(\textbf{x}_i, s(\textbf{x}_i)) \big| | s(\textbf{x}_i) | < \tau\}_{i=1}^{| V |}, 
\label{eq: sparse_voxel}
\end{equation}
where $s(\textbf{x}_i)$ denotes the SDF value at position $\textbf{x}_i$. 

\vspace{12pt}
\subsection{Symmetric Network Architecture} 
Our fully end-to-end SDF VAE framework adopts a symmetric encoder-decoder network architecture, as illustrated in the upper half of Figure~\ref{fig:pipeline}. Specifically, the encoder employs a hybrid framework combining sparse 3D convolution networks and transformer networks. We first extract local geometric features through a series of residual sparse 3D CNN blocks interleaved with 3D mean pooling operations, progressively downsampling the spatial resolution. We then process the sparse voxels as variable-length tokens and utilize 
shifted window attention to capture local contextual information between the valid voxels.
Inspired by Trellis~\cite{xiang2024structured}, the feature of each valid voxel is augmented with positional encoding based on its 3D coordinates before being fed into 3D shift window attention layers. This hybrid design outputs sparse latent representations at reduced resolution $(\frac{R}{f})^3$, where $f$ denotes the downsampling factor. The decoder of our SS-VAE adopts a symmetric structure with respect to the encoder, leveraging attention layers and sparse 3D CNN blocks to progressively upsample the latent representation and reconstruct the SDF volume $\tilde{V}$.

\subsection{Training Losses} 
The decoded sparse voxels $\tilde{V}$ contain both the input voxels $\tilde{V}_\text{in}$ and additional valid voxels $\tilde{V}_\text{extra}$. We enforce supervision on the SDF values across all these spatial positions. To enhance geometric fidelity, we impose additional supervision on the active voxels situated near the sharp edges of the mesh, specifically in regions exhibiting high-curvature variations on the mesh surface.
Moreover, the term of KL-divergence regularization is imposed on the latent representation $\mathbf{z}$ to constrain excessive variations in the latent space. The overall training objective of our SS-VAE is formulated as:
\vspace{-4pt}
\begin{equation}
    \mathcal{L}_{c} =
  \frac{1}{\lvert\tilde{V}_{c}\rvert}
  \sum_{(\mathbf{x},\,\tilde{s}(\mathbf{x}))\in\tilde{V}_{c}}
  \!\bigl\lVert s(\mathbf{x})-\tilde{s}(\mathbf{x})\bigr\rVert_2^{2},
  \quad
  c\in\{\text{in},\text{ext},\text{sharp}\},
\end{equation}
\vspace{-6pt}
\begin{equation}
    \mathcal{L}_{\text{total}} =
  \sum_{c}\lambda_{c}\,\mathcal{L}_{c} +
  \lambda_{\text{KL}}\,\mathcal{L}_{\text{KL}},
\end{equation}
\vspace{-4pt}
where $\lambda_\text{in}$, $\lambda_\text{ext}$, $\lambda_\text{sharp}$ and $\lambda_\text{KL}$ denote the weight of each term.

\subsection{Multi-resolution Training} 
To enhance training efficiency and enable our SS-VAE to encode meshes across varying resolutions, we utilize the multi-resolution training paradigm. Specifically, during each training iteration, we randomly 
sample a target resolution from the candidate set $\{256^3, 384^3, 512^3, 1024^3\}$, then trilinearly interpolate the input SDF volume to the selected resolution before feeding it into the SS-VAE. 

\section{Spatial Sparse Attention and DiT}
\label{dit}

\begin{figure*}[!t]
  \centering
  \includegraphics[width=0.9\linewidth]{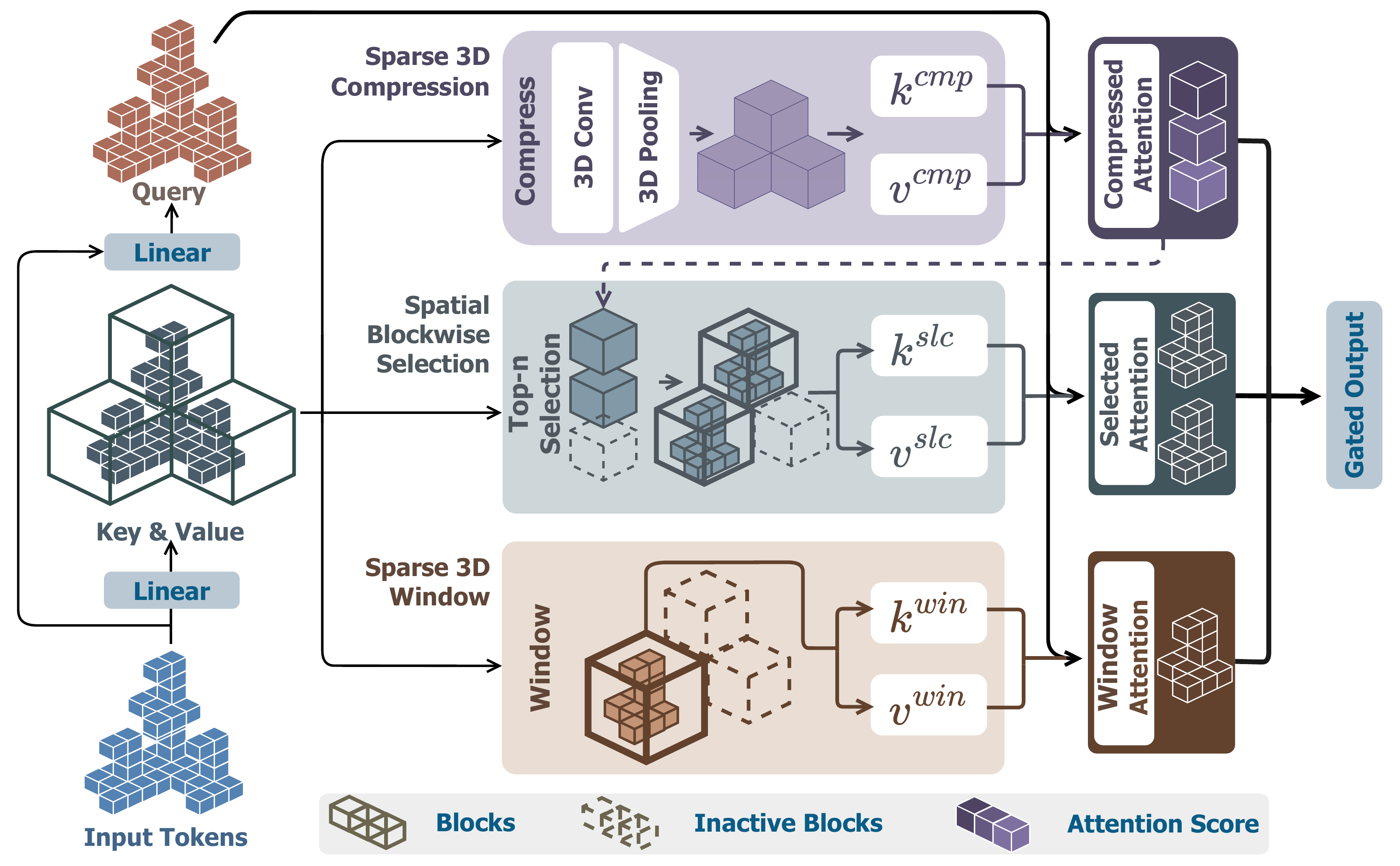}
  \caption{The framework of our Spatial Sparse Attention (SSA). We partition the input tokens into blocks based on their 3D coordinates, and then construct key-value pairs through three distinct modules. For each query token, we utilize sparse 3D compression module to capture global information, while the spatial blockwise selection module selects important blocks based on compression attention scores to extract fine-grained features, and the sparse 3D window module injects local features. Ultimately, we aggregate the final output of SSA from the three modules using predicted gate scores.}
  \label{fig:ssa}
\end{figure*}

Through our SS-VAE framework, 3D shapes can be encoded into latent representations $\mathbf{z}$. Following a methodology analogous to Trellis~\cite{xiang2024structured}, we serialize the latent tokens $\mathbf{z}$ and train a rectified flow transformer-based 3D shape generator conditioned on input images. To ensure efficient generation of high-resolution meshes, we propose spatial sparse attention that substantially accelerates both training and inference processes. Furthermore, we introduce a sparse conditioning mechanism to extract the foreground region of the input images, thereby reducing the number of conditioning tokens. The architecture of the DiT is illustrated in the lower half of Figure~\ref{fig:pipeline}.

\subsection{Spatial Sparse Attention}
Given input tokens $\mathbf{q}$, $\mathbf{k}$, $\mathbf{v}$ $\in \mathbb{R}^{N\times d}$, where $N$ denotes the token length, and $d$ represents the head dimension, the standard full attention is formulated as:
\begin{align}
\mathbf{o}_t
  &= \text{Attn}\!\bigl(\mathbf{q}_t, \mathbf{k}, \mathbf{v}\bigr) \nonumber\\
  &= \sum_{i=1}^{N}
     \frac{\mathbf{p}_{t,i}\,\mathbf{v}_i}
          {\displaystyle\sum_{j=1}^{N}\mathbf{p}_{t,j}},
     \quad t\in[0,N), \label{eq:attn}\\[4pt]
\mathbf{p}_{t,j}
  &= \exp\!\left(\frac{\mathbf{q}_t^{\top}\mathbf{k}_j}{\sqrt{d}}\right).
\end{align}
As the resolution of SS-VAE escalates, the length of input tokens grows substantially, reaching over 100k at a resolution of $1024^3$, leading to prohibitively low computational efficiency in attention operations. Inspired by NSA (Native Sparse Attention)~\cite{yuan2025native}, we proposes Spatial Sparse Attention mechanism, which partitions key and value tokens into spatially coherent blocks based on their geometric relationships and performs blockwise token selection to achieve significant acceleration.

A naive implementation involves treating latent tokens $\mathbf{z}$ as a 1D sequence and partitioning it into fixed-length blocks based on token indices, analogous to NSA. However, this approach suffers from two critical limitations: On the one hand, tokens within the same block may not be spatially adjacent in 3D space, despite sharing contiguous indices. On the other hand, due to the sparse voxel structure, blocks with identical indices across different samples occupy divergent spatial regions. These issues collectively lead to unstable training convergence. To resolve these challenges, we propose partitioning blocks based on 3D coordinates. As illustrated in Figure~\ref{fig:ssa}, we divide the 3D space into subgrids of size $m^3$, where active tokens from sparse voxels residing in the same subgrid are grouped into one block. Our Spatial Sparse Attention comprises three core modules: sparse 3D compression, spatial blockwise selection, and sparse 3D window. The attention computation proceeds as follows:
\begin{equation}
\label{gate}
\begin{aligned}
    \mathbf{o}_t &=
        \omega^\text{cmp}_t \text{Attn}(\mathbf{q}_t, \mathbf{k}_t^\text{cmp}, \mathbf{v}_t^\text{cmp}) \\
        &\quad + \omega^\text{slc}_t \text{Attn}(\mathbf{q}_t, \mathbf{k}_t^\text{slc}, \mathbf{v}_t^\text{slc}) \\
        &\quad + \omega^\text{win}_t \text{Attn}(\mathbf{q}_t, \mathbf{k}_t^\text{win}, \mathbf{v}_t^\text{win}),
\end{aligned}
\end{equation}
where $\mathbf{k}_t$ and $\mathbf{v}_t$ represent the selected key and value tokens in each module for query $\textbf{q}_t$, respectively. $\omega_t$ is the gating score for each module, obtained by applying a linear layer followed by a sigmoid activation to the input features.

\vspace{12pt}
\noindent\textbf{Sparse 3D Compression.}
After partitioning input tokens into spatially coherent blocks based on their 3D coordinates, we leverage a compression module to extract block-level representations of the input tokens. Specifically, we first incorporate intra-block positional encoding for each token within a block of size $m_\text{cmp}^3$, then employ sparse 3D convolution followed by sparse 3D mean pooling to compress the entire block:
\begin{equation}
\label{cmp}
    \mathbf{k}^\text{cmp}_t = \delta(\mathbf{k}_t + \text{PE}(\mathbf{k}_t)),
\end{equation}
where $\mathbf{k}^\text{cmp}_t$ denotes the block-level key token, $\text{PE}(\cdot)$ is absolute position encoding, and $\delta(\cdot)$ represents operations of sparse 3D convolution and sparse 3D mean pooling. The sparse 3D compression module effectively captures block-level global information while reducing the number of tokens, thereby enhancing computational efficiency.

\vspace{12pt}
\noindent\textbf{Spatial Blockwise Selection.} 
The block-level representations only contain coarse-grained information, necessitating the retention of token-level features to enhance the fine details in the generated 3D shapes. However, the excessive number of input tokens leads to computationally inefficient operations if all tokens are utilized. By leveraging the sparse 3D compression module, we compute the attention scores $\mathbf{s}_\text{cmp}$ between the query $\mathbf{q}$ and each compression block, subsequently selecting all tokens within the top-$k$ blocks exhibiting the highest scores. The resolution $m_\text{slc}$ of the selection blocks must be both greater than and divisible by the resolution $m_\text{cmp}$ of the compression blocks. The relevance score $\mathbf{s}^\text{slc}_t$ for a selection block is aggregated from its constituent compression blocks. GQA (Grouped-Query Attention)~\cite{ainslie2023gqa} is employed to further improve computational efficiency, the attention scores of the shared query heads within each group are accumulated as follows:
\begin{equation}
\label{eq:score}
    \mathbf{s}^\text{slc}_t = \sum_{i \in \mathcal{B}_{\text{cmp}}} \sum_{h=1}^{h_s} s^{\text{cmp}, i}_{t, h},
\end{equation}
where $\mathcal{B}_{\text{cmp}}$ denotes the set of compression blocks within the selection block, and $h_s$ represents the number of shared heads within a group.
The top-$k$ selection blocks with the highest $\mathbf{s}^\text{slc}_t$ scores are selected, and all tokens contained within them are concatenated to form $\mathbf{k}^\text{slc}_t$ and $\mathbf{v}^\text{slc}_t$, which are used to compute the spatial blockwise selection attention. 

We implement the spatial blockwise selection attention kernel using Triton~\cite{tillet2019triton}, with
two key challenges arising from the sparse 3D voxel structures: 1) the number of tokens varies across different blocks, and 2) tokens within the same block may not be contiguous in HBM. To address these, we first sort the input tokens based on their block indices, then compute the starting index $\mathcal{C}$ of each block as kernel input. In the inner loop, $\mathcal{C}$ dynamically governs the loading of corresponding block tokens. The complete procedure of forward pass is formalized in Algorithm~\ref{alg:slc}.

\input{tables/algo_ssa}
\vspace{12pt}
\noindent\textbf{Sparse 3D Window.} 
In addition to sparse 3D compression and spatial blockwise selection modules, we further employ an auxiliary sparse 3D window module to explicitly incorporate localized feature interactions. Drawing inspiration from Trellis~\cite{xiang2024structured}, we partition the input token-containing voxels into non-overlapping windows of size $m_\text{win}^3$. For each token, we formulate its contextual computation by dynamically aggregating active tokens within the corresponding window to form $\mathbf{k}^\text{win}_t$ and $\mathbf{v}^\text{win}_t$, followed by localized self-attention calculation exclusively over this constructed token subset.

Through the modules of sparse 3D compression, spatial blockwise selection, and sparse 3D window, corresponding key-value pairs are constructed. Subsequently, attention calculations are performed for each module, and the results are aggregated and weighted according to gate scores to produce the final output of the spatial sparse attention mechanism.

\subsection{Sparse Conditioning Mechanism}
Existing image-to-3D models~\cite{li2024craftsman, wu2024direct3d, zhang2024clay} typically employ DINO-v2~\cite{oquab2023dinov2} to extract pixel-level features from conditional images, followed by cross-attention operation with noisy tokens to achieve conditional generation. However, for a majority of input images, more than half of the regions consist of background, which not only introduces additional computational overhead but may also adversely affect the alignment between the generated meshes and the conditional images. To mitigate this issue, we propose a sparse conditioning mechanism that selectively extracts and processes sparse foreground tokens from input images for cross-attention computation. Formally, given an input image $\mathcal{I}$, the sparse conditioning tokens $\mathbf{c}$ are computed as follows:
\begin{equation}
    \mathbf{c} = \text{Linear}(f(E_\text{DINO}(\mathcal{I}))) + \text{PE}(f(E_\text{DINO}(\mathcal{I}))),
\end{equation}
where $E_\text{DINO}$ is the DINO-v2 encoder, $f(\cdot)$ denotes the operation of extracting the foreground tokens based on the mask, $\text{PE}(\cdot)$ is the absolute position encoding, and $\text{Linear}(\cdot)$ represents a linear layer. Then we perform cross attention using the finalized sparse conditioning tokens $\mathbf{c}$ and the noisy tokens.

\subsection{Rectified Flow}

We employ rectified flow objective~\cite{esser2024scaling, lipman2022flow} to train our generative model. Rectified flow defines forward process as linear trajectory between data distribution and standard normal distribution:
\begin{equation}
\label{eq:rf_forward}
    \mathbf{x}(t) = (1 - t)\mathbf{x}_0 + t \epsilon,
\end{equation}
where $\epsilon$ is the noise, and $t$ denotes the timestep. Our generative model is trained to predict the velocity field from noisy samples to the data distribution. The training loss is formulated using conditional flow matching, formulated as follows:
\begin{equation}
\label{eq:rf_loss}
    \mathcal{L}_\text{CFM} = \mathbb{E}_{t, \mathbf{x}_0, \epsilon}\|\mathbf{v}_\theta(\mathbf{x}_t, \mathbf{c}, t) - (\epsilon - \mathbf{x}_0)\|^2_2,
\end{equation}
where $\mathbf{v}_\theta$ is the neural networks.

%% file: tables/algo_ssa.tex
\begin{algorithm*}[!h]
\centering
  \caption{\small\label{alg:slc}Spatial Blockwise Selection Attention Forward Pass}
  \begin{algorithmic}[1]
    \REQUIRE $\mathbf{q} \in \mathbb{R}^{N\times (h_{kv} \times h_s) \times d}$, $\mathbf{k}\in \mathbb{R}^{N\times h_{kv} \times d}$ and $\mathbf{v} \in \mathbb{R}^{N\times h_{kv} \times d}$, number of key/value heads $h_{kv}$, number of the shared heads $h_s$, number of the selected blocks $T$, indices of the selected blocks $\mathbf{I} \in \mathbb{R}^{N\times h_{kv} \times T}$, the number of divided key/value blocks $N_b$, index of the first token in each block $\mathcal{C}\in\mathbf{R}^{N_b+1}$, block size $B_k$.

    \STATE Divide the output $\mathbf{o}\in \mathbb{R}^{N\times (h_{kv} \times h_s) \times d}$ into $(N, h_{kv})$ blocks, each of size $h_s \times d$. Divide the logsumexp $l \in \mathbb{R}^{N\times (h_{kv} \times h_s)}$ into $(N, h_{kv})$ blocks, each of size $h_s$.
    
    \STATE Sort all tokens within $\mathbf{q}$, $\mathbf{k}$ and $\mathbf{v}$ according to their respective block indices. 

    \FOR{$t = 1$ \textbf{to} $N$}
        \FOR{$h = 1$ \textbf{to} $h_{kv}$}
            \STATE
            Initialize $\mathbf{o}_{t,h}=(0)_{h_s \times d} \in\mathbb{R}^{h_s \times d}$, logsumexp $l_{t,h}=(0)_{h_s}\in\mathbb{R}^{h_s}$, and $\mathbf{m}_{t,h}=(-\text{inf})_{h_s}\in\mathbb{R}^{h_s}$.
            
            \STATE Load $\mathbf{q}_{t,h}\in\mathbb{R}^{h_s \times d}$, $\mathbf{I}_{t,h} \in \mathbb{R}^T$ from HBM to on-chip SRAM.
            
            \FOR{$j = 1$ \textbf{to} $T$}
                \STATE Load the starting token index $b_s = \mathcal{C}^ {(\mathbf{I}_{t,h}^{(j)})}$ and ending token index $b_e = \mathcal{C}^ {(\mathbf{I}_{t,h}^{(j)})+1} - 1$ of the $\mathbf{I}_{t,h}^{(j)} th$ block from HBM to on-chip SRAM.
                \FOR{$i = b_s$ \textbf{to} $b_e$ \textbf{by} $B_k$}
                    \STATE Load $\mathbf{k}_i$, $\mathbf{v}_i \in \mathbb{R}^{B_k \times d}$ from HBM to on-chip SRAM.
                    \STATE Compute $\mathbf{s}^{(i)}_{t,h}=\mathbf{q}_{t,h} \mathbf{k}_i^T \in \mathbb{R}^{h_s\times B_k}$.
                    
                    \STATE Compute $\mathbf{m}_{t,h}^{(i)}=\text{max}(\mathbf{m}_{t,h}, \text{rowmax}(\mathbf{s}^{(i)}_{t,h}))\in\mathbb{R}^{h_s}$.
                    
                    \STATE Compute $\mathbf{p}_{t,h}^{(i)}=e^{\mathbf{s}_{t,h}^{(i)}-\mathbf{m}_{t,h}^{(i)}}\in\mathbb{R}^{h_s\times B_k}$.
                    
                    \STATE Compute $\mathbf{o}_{t,h} = e^{\mathbf{m}_{t,h}-\mathbf{m}_{t,h}^{(i)}} \mathbf{o}_{t,h} + \mathbf{p}_{t,h}^{(i)} \mathbf{v}_i$.
                    
                    \STATE Compute $l_{t,h}=\mathbf{m}_{t,h}^{(i)}+\text{log}(e^{l_{t,h}-\mathbf{m}_{t,h}^{(i)}} + \text{rowsum}(\mathbf{p}_{t,h}^{(i)}))$, $\mathbf{m}_{t,h}=\mathbf{m}_{t,h}^{(i)}$.
                \ENDFOR
            \ENDFOR
            
            \STATE Compute $\mathbf{o}_{t,h}=e^{\mathbf{m}_{t,h}-l_{t,h}}\mathbf{o}_{t,h}$.

            \STATE Write $\mathbf{o}_{t,h}$ and $l_{t,h}$ to HBM as the $(t, h)$-th block of $\mathbf{o}$ and $l$, respectively.
        \ENDFOR
    \ENDFOR  
    \STATE Return the output $\mathbf{o}$ and the logsumexp $l$.

  \end{algorithmic}
\end{algorithm*}

%% file: contents/4_experiments.tex
\vspace{16pt}
\section{Experiments}
\label{sec:experiments}

\vspace{8pt}
\subsection{Datasets}
Our Direct3D-S2 is trained on publicly available 3D datasets including Objaverse~\cite{deitke2023objaverse}, Objaverse-XL~\cite{deitke2023objaversexl}, and ShapeNet~\cite{chang2015shapenet}. Due to the prevalence of low-quality meshes in these collections, we curated approximately 452k 3D assets through rigorous filtering for training. Following prior approach~\cite{zhang2024clay} in geometry processing, we first convert the original non-watertight meshes into watertight ones, then compute ground-truth SDF volumes that serve as both input to and supervision for our SS-VAE. For training our image-conditioned DiT, we render 45 RGB images per mesh at $1024\times1024$ resolution with random camera parameters. The camera configuration space is defined as follows: elevation angles ranging from $10^\circ$ to $40^\circ$, azimuth angles spanning $[0^\circ, 180^\circ]$, and focal lengths varying between 30mm and 100mm. To rigorously evaluate the geometric fidelity of meshes generated by Direct3D-S2, we established a challenging benchmark comprising highly detailed images sourced from professional communities including Neural4D~\cite{neural4d}, Meshy~\cite{meshy}, and CivitAI~\cite{civitai}. The quantitative assessment employs ULIP-2~\cite{xue2024ulip}, Uni3D~\cite{zhou2023uni3d} and OpenShape~\cite{liu2023openshape} metrics to measure shape-image alignment between generated meshes and conditional input images, enabling systematic comparison with state-of-the-art 3D generation methods. 

\begin{figure*}
  \centering
  \includegraphics[width=1\linewidth]{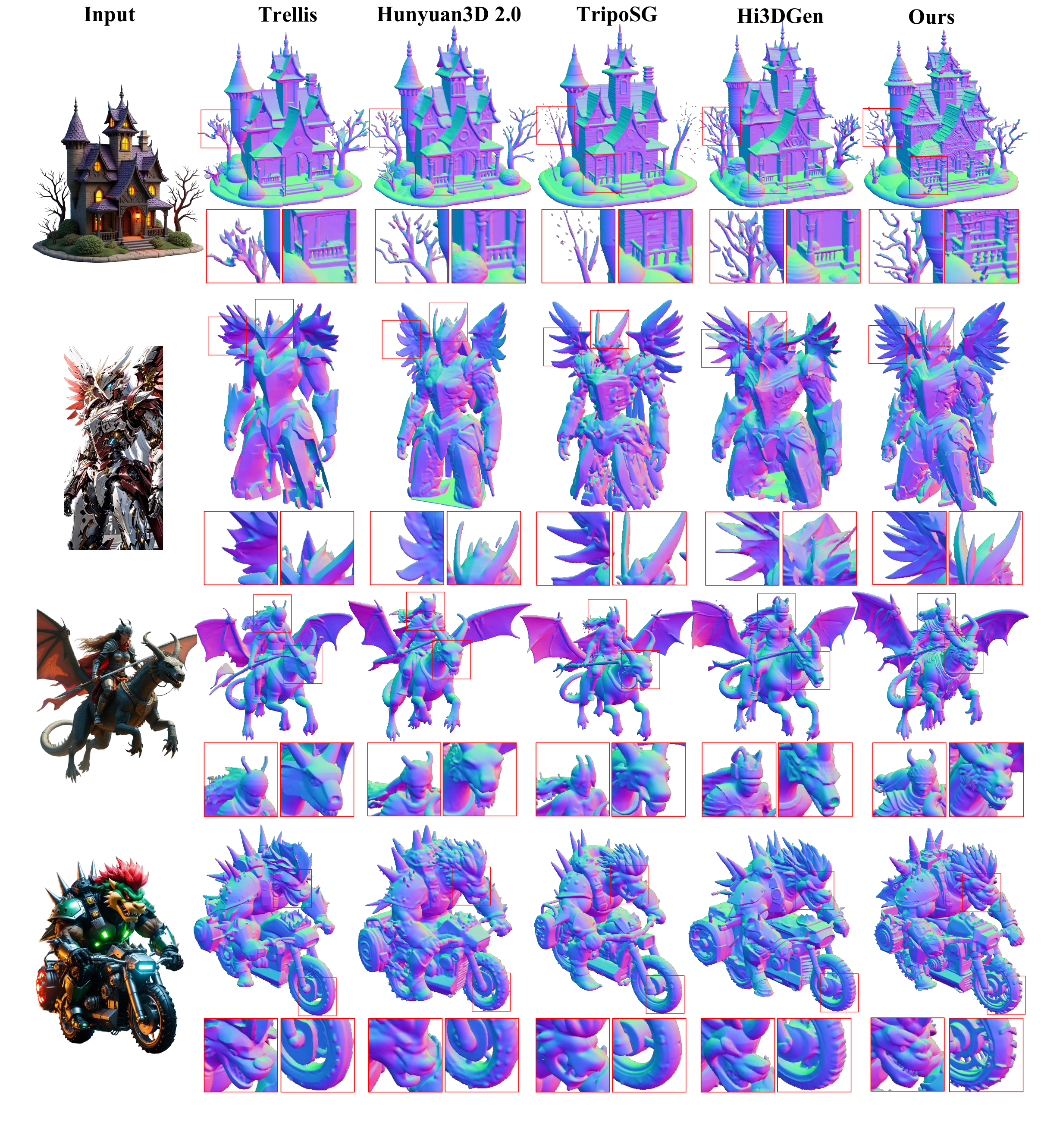}
  \caption{Qualitative comparisons between other image-to-3D methods and our approach.}
  \label{fig:dit_compare1}
\end{figure*}

\subsection{Implementation Details}
\noindent \textbf{VAE.} 
We utilize active voxels from volumes with SDF values less than $\tau=\frac{1}{128}$ as inputs to the SS-VAE. The downsampling factor $f$ for the encoder is set to 8, and the channel dimension of the latent representation $\mathbf{z}$ is configured to 16. The weights for the various losses are set as: $\lambda_\text{in}=1.0$, $\lambda_\text{ext}=1e-1$, $\lambda_\text{sharp}=1.0$, and $\lambda_\text{KL}=1e-3$. We employ the AdamW~\cite{loshchilov2017decoupled} optimizer with an initial learning rate of $1e-4$. To enhance training efficiency, we first conduct multi-resolution training using SDF volumes at three resolutions of $\{256^3, 384^3, 512^3\}$ over a period of one day on 8 A100 GPUs, with a batch size of 4 per GPU. Subsequently, we fine-tune the SS-VAE for one additional day at a resolution of $1024^3$ with a learning rate of $1e-5$ with a batch size of 1 per GPU.
\vspace{+2pt}

\noindent \textbf{DiT.}
\vspace{+2pt}
Our SS-DiT comprises 24 layers of DiT blocks with a hidden dimension of 1024. We employ Grouped-Query Attention (GQA)~\cite{ainslie2023gqa} with a group number set to 2, where each group contains 16 attention heads. The hidden dimension of each head is configured as 32. For the spatial sparse attention (SSA) mechanism, we configure the resolution of the compression blocks to $m_\text{cmp}=4$ , the resolution of the selection blocks to $m_\text{slc}=8$, and the size of the sparse 3D windows $m_\text{win}=8$. We utilize DINO-v2 Large~\cite{oquab2023dinov2} to extract features from conditional images, with input images having a resolution of $518\times518$. For the DiT, we implement a progressive training strategy that gradually increases the resolution from $256^3$ to $1024^3$ to accelerate convergence. Table~\ref{tab:dit_config} presents the average number of latent tokens, learning rate, batch size, and training duration settings at different resolutions. We employ the AdamW optimizer and trained the model for a total of 7 days on 8 A100 GPUs. For the $1024^3$ resolution, we further filtered 68k high-fidelity 3D assets for training. Additionally, similar to Trellis~\cite{xiang2024structured}, we trained an extra DiT to predict the indices of the sparse latent tokens $\mathbf{z}$, which took 7 days on 8 A100 GPUs.

\input{tables/dit_config}
\subsection{Quantitative and Qualitative Comparisons}

\input{tables/compare_dit}

To empirically validate the effectiveness of our Direct3D-S2 framework, we conduct comprehensive experiments against state-of-the-art image-to-3D approaches. Our systematic evaluation employs three multimodal models: ULIP-2~\cite{xue2024ulip}, Uni3D~\cite{zhou2023uni3d}, and OpenShape~\cite{liu2023openshape}, to assess the similarity between the generated meshes and input images. The quantitative results are reported in Table~\ref{tab:comparison_i23d}, where it is evident that our Direct3D-S2 outperforms the other approaches across three metrics, indicating that the meshes produced by our Direct3D-S2 achieve better alignment with the input images. Moreover, we present qualitative comparisons in Figure~\ref{fig:dit_compare1}. Although the other methods generate overall satisfactory results, they struggle to capture finer structures due to resolution limitations, as illustrated by 
the railings of the house and surrounding branches of trees in the first row. In contrast, thanks to our proposed SSA mechanism, our Direct3D-S2 is capable of generating high-resolution meshes, achieving superior quality even for these intricate details. We provide more qualitative comparisons with both open-source and closed-source approaches in Figure~\ref{fig:mesh_compare}.

\begin{figure}[!t]
  \centering
  \includegraphics[width=1\linewidth]{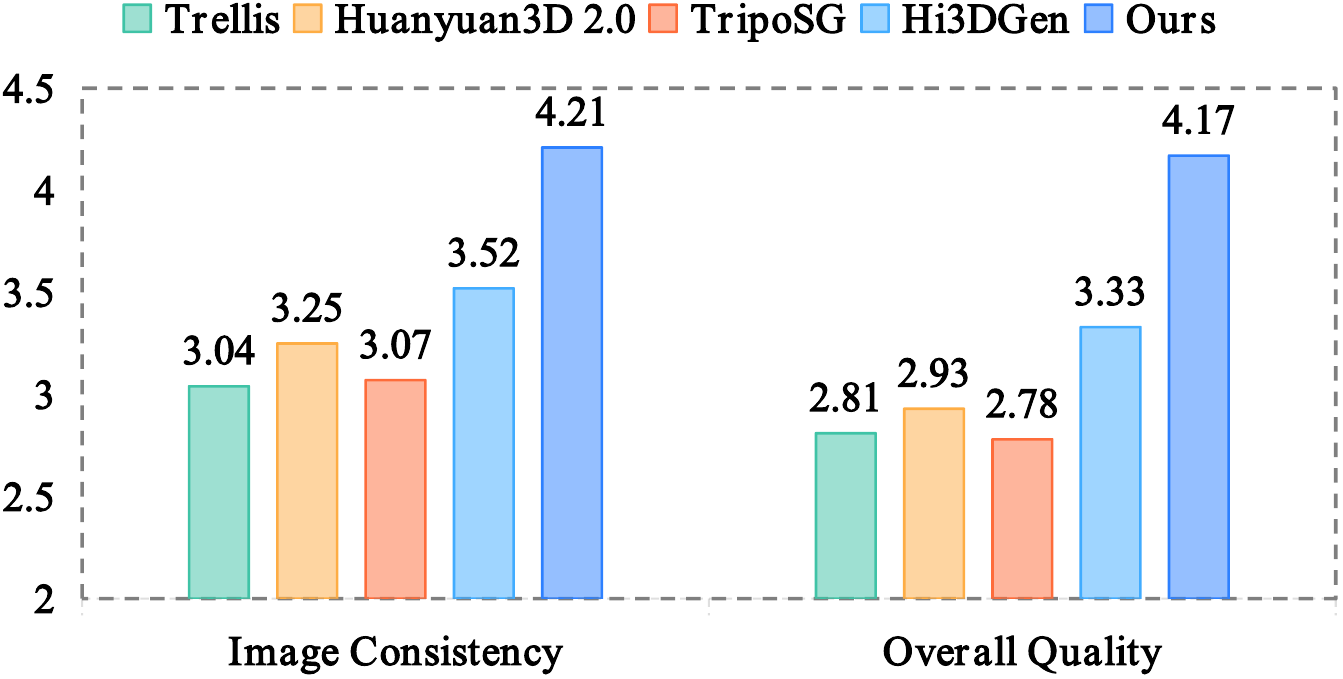}
  \caption{User Study for Image-to-3D Generation.}
  \label{fig:user_study}
\end{figure}
In addition, we conducted a user study with 40 participants evaluating 75 unfiltered meshes generated by our Direct3D-S2 and other image-to-3D methods. Participants scored each output using two criteria: image consistency and overall geometric quality, with scores ranging from 1 (poorest) to 5 (excellent). As shown in Figure~\ref{fig:user_study}, Our Direct3D-S2 demonstrates statistically superiority over other approaches across both evaluation metrics.

\section{Comparison of VAE}

\begin{figure*}[!t]
  \centering
  \includegraphics[width=1.0\linewidth]{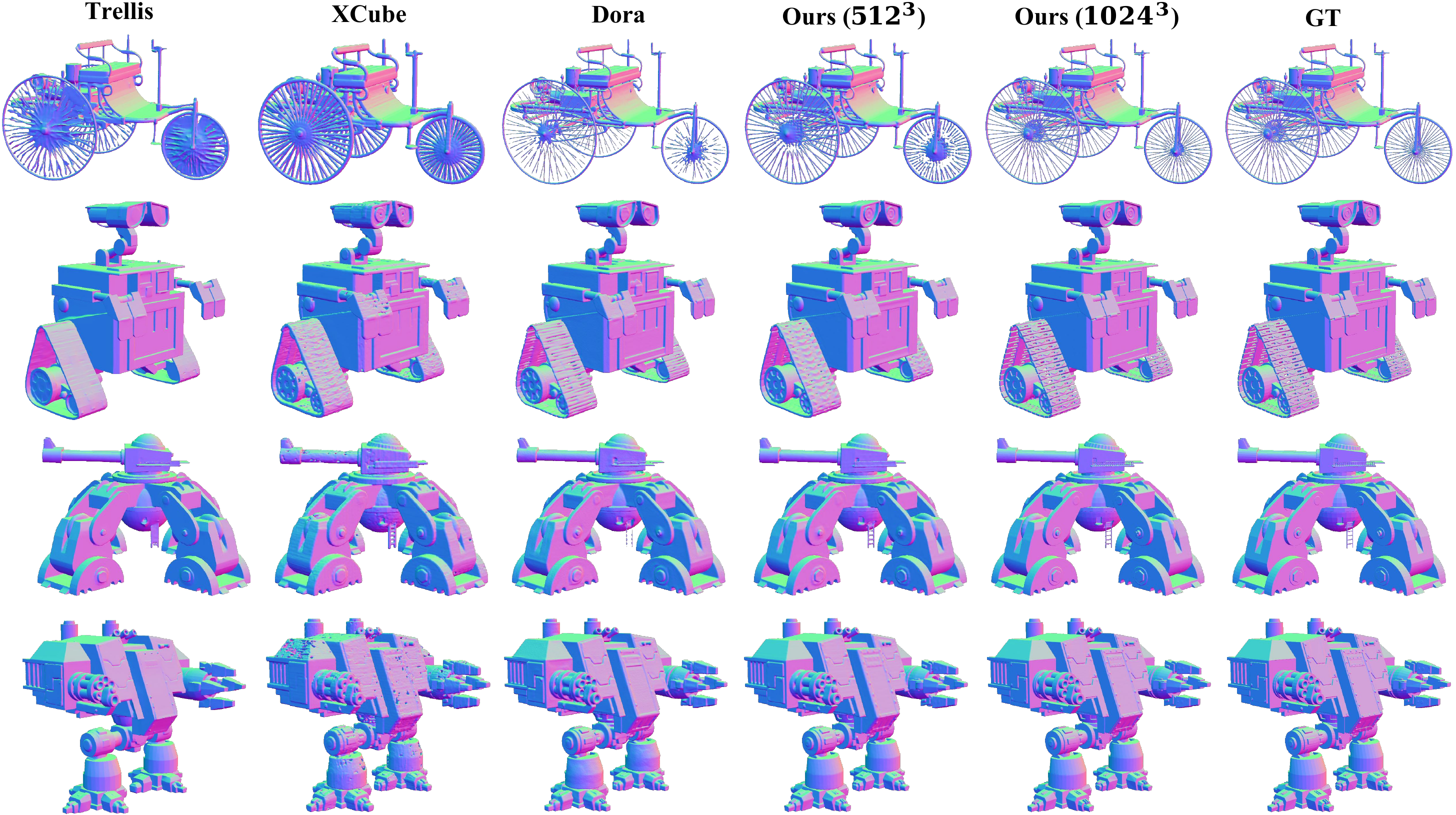}
  \caption{Qualitative comparisons of VAE reconstruction results. Note that we used a latent token length of 4096 during the inference of Dora~\cite{chen2024dora}.}
  \label{fig:compare_vae}
\end{figure*}

To validate the reconstruction quality of our SS-VAE, we curated a challenging validation set from the Objaverse~\cite{deitke2023objaverse} dataset, comprising meshes with complex geometric structures. Qualitative comparisons with competing methods are shown in Figure~\ref{fig:compare_vae}. It can be observed our SS-VAE achieves superior reconstruction accuracy at $512^3$ resolution, and demonstrates markedly improved performance on complex geometries at $1024^3$ resolution. Notably, thanks to our fully end-to-end SDF reconstruction framework, SS-VAE requires only 2 days of training on 8 A100 GPUs, significantly fewer than competing methods that typically demand at least 32 GPUs for equivalent training durations.

\begin{figure*}[!t]
  \centering
  \includegraphics[width=1.\linewidth]{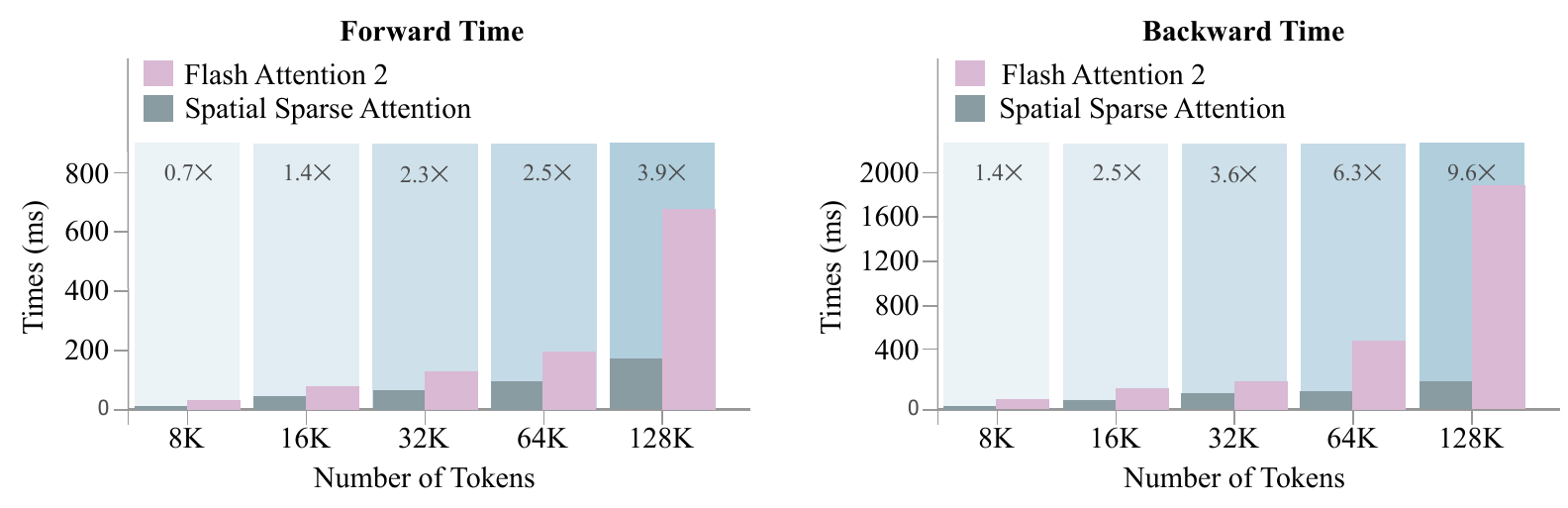}
  \caption{Comparison of the forward and backward time of our SSA and FlashAttention-2.}
  \label{fig:speed}
\end{figure*}

\begin{figure*}[!t]
  \centering
  \includegraphics[width=0.9\linewidth]{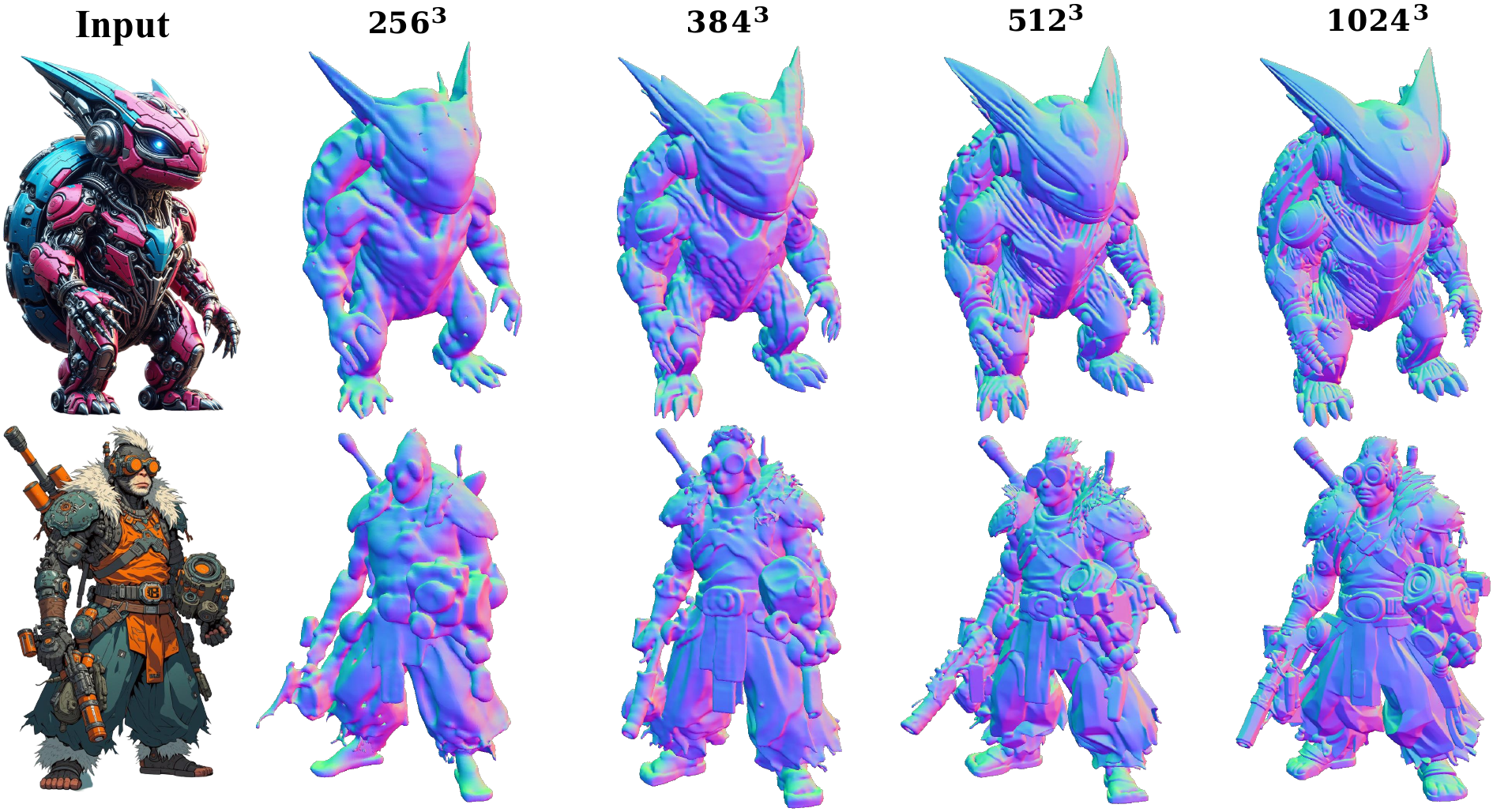}
  \caption{The visualization results of our Direct3D-S2 for image-to-3D generation across four resolutions: \{$256^3$, $384^3$, $512^3$, $1024^3$\}.}
  \label{fig:ablation_res}
\end{figure*}

\begin{figure*}[!h]
  \centering
  \includegraphics[width=0.9\linewidth]{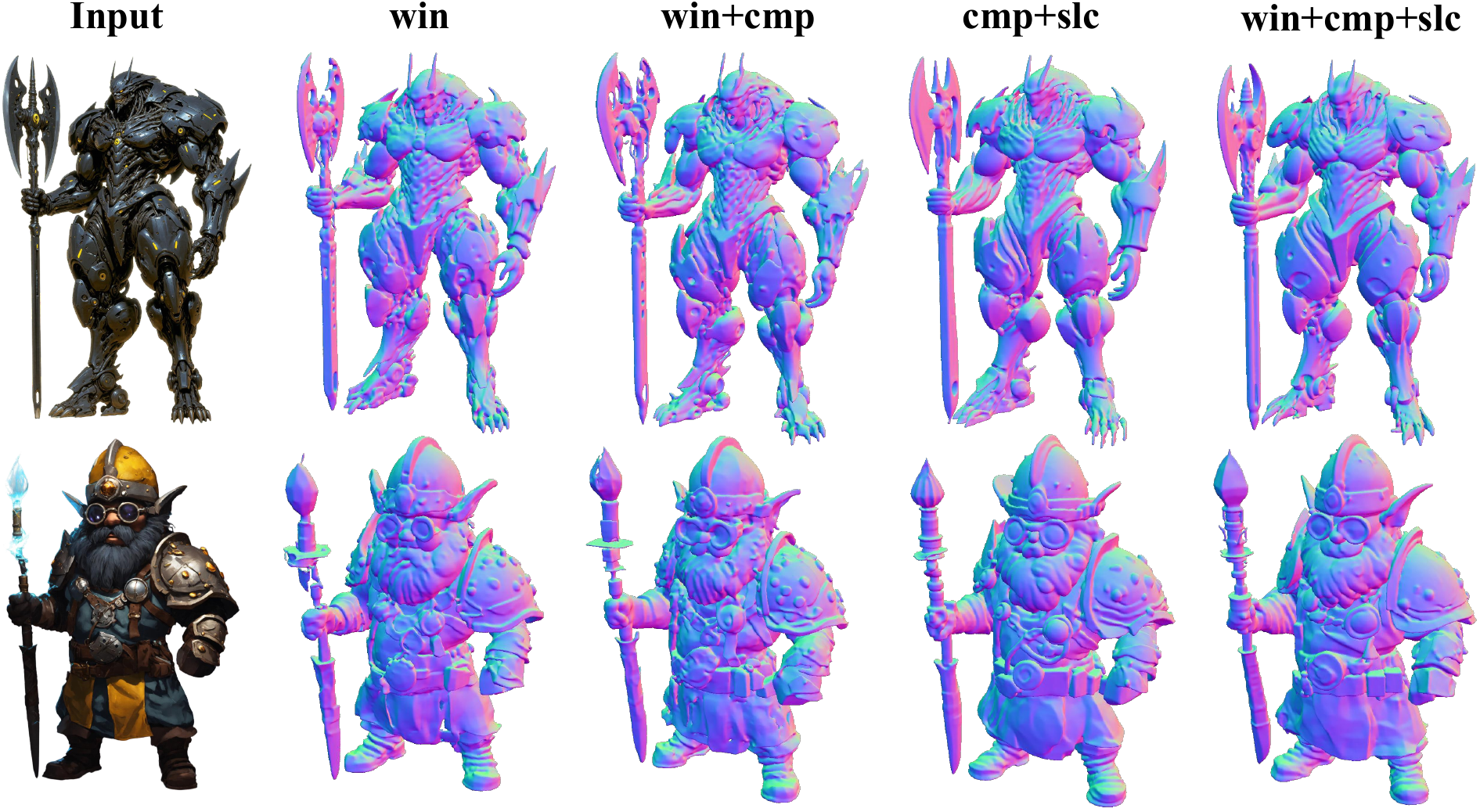}
  \caption{Ablation studies for the three modules of SSA at resolution $512^3$, where \emph{win}, \emph{cmp}, and \emph{slc} denote the sparse 3D window, sparse 3D compression, and spatial blockwise selection modules, respectively.}
  \label{fig:ablation_module}
\end{figure*}

\begin{figure}[!t]
  \centering
  \includegraphics[width=1\linewidth]{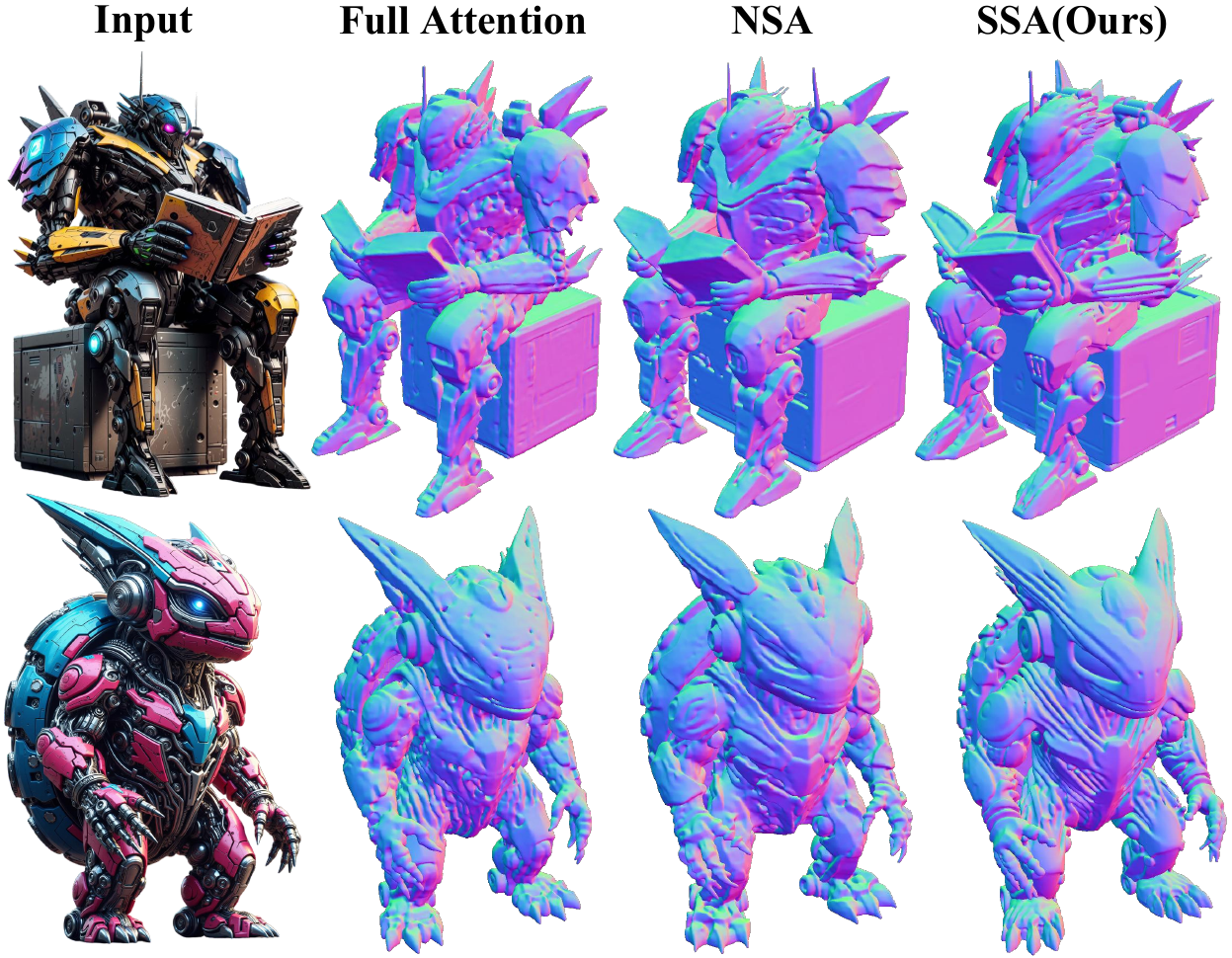}
  \caption{Ablation studies of our proposed SSA mechanism.}
  \label{fig:ablation_nsa}
  \vspace{-4pt}
\end{figure}

\begin{figure}[!t]
  \centering
  \includegraphics[width=1\linewidth]{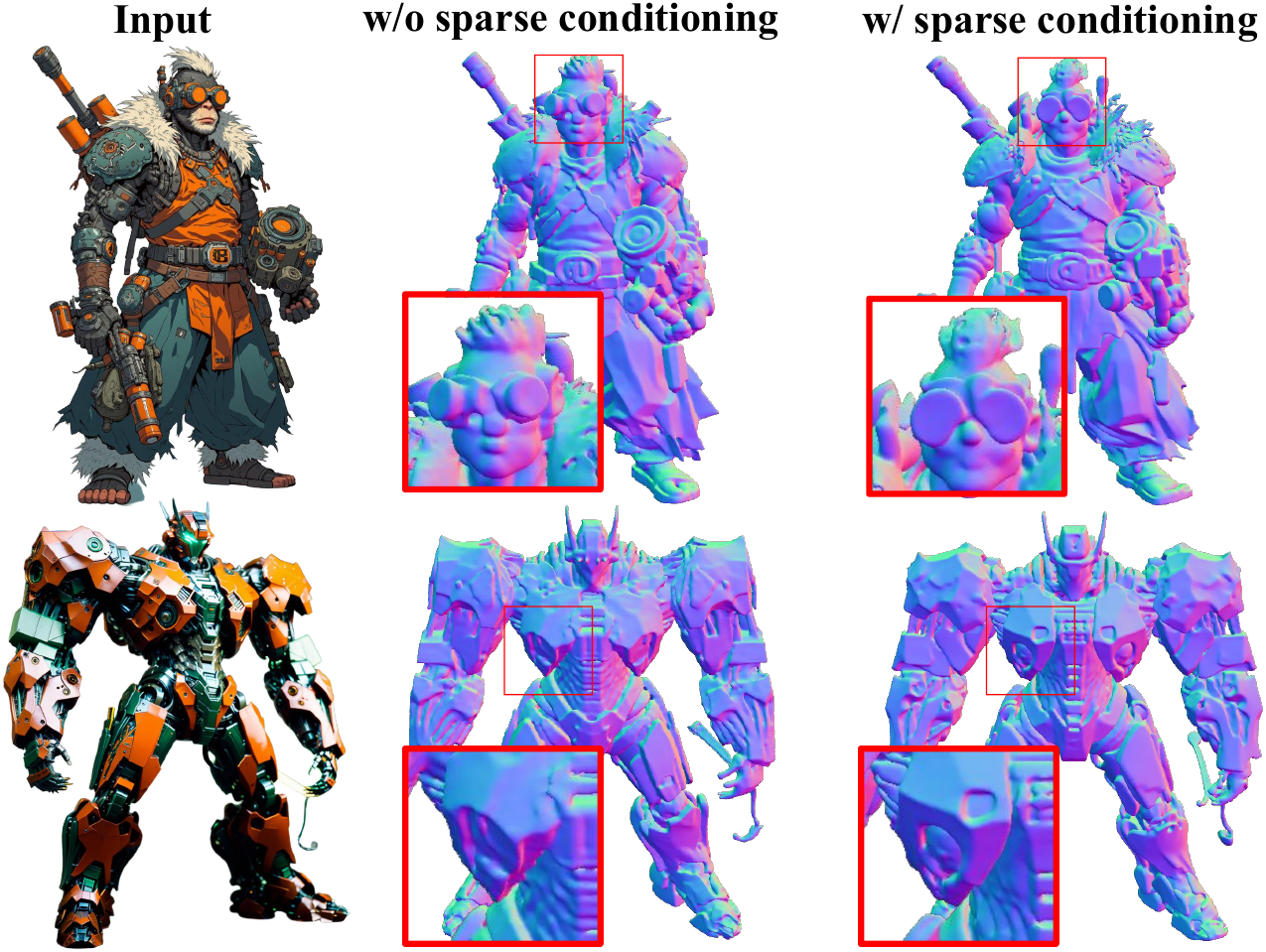}
  \caption{Ablation studies for sparse conditioning mechanism.}
  \label{fig:ablation_condition}
  \vspace{-4pt}
\end{figure}

\subsection{Ablation Studies}
\noindent\textbf{Image-to-3D Generation in Different Resolution. }
We present the generation results of our Direct3D-S2 across four resolutions \{$256^3, 384^3, 512^3, 1024^3$\} in Figure~\ref{fig:ablation_res}. The results demonstrate that increasing resolution progressively improves mesh quality. At lower resolutions $256^3$ and $384^3$, the generated meshes exhibit limited geometric details and misalignment with input images. At $512^3$ resolution, the meshes display enhanced high-frequency geometric details. Further increasing the resolution to $1024^3$ yields meshes with sharper edges and improved alignment with input image details.

\vspace{12pt}
\noindent\textbf{Effect of Each Module in SSA.}
We validated the effect of the three modules in SSA at resolution $512^3$, with the results presented in Figure~\ref{fig:ablation_module}. When using only the sparse 3D window module (\emph{win}), the generated meshes exhibited detailed structures but suffered from surface irregularities due to the lack of global context modeling. Introducing the sparse 3D compression module (\emph{win+cmp}) showed minimal performance changes, which is reasonable as this module primarily serves to obtain the attention scores for the blocks. After incorporating the spatial blockwise selection module (\emph{win+cmp+slc}), the model can focus on the most important regions globally, resulting in a notable improvement in mesh quality. We also observed that not utilizing the window (\emph{cmp+slc}) did not result in a significant drop in model performance, but slowed convergence, demonstrating that local feature interaction contributes to more stable training and enhances convergence speed.

\vspace{12pt}
\noindent \textbf{Runtime of Different Attention Mechanisms.} 
We implemented a custom Triton~\cite{tillet2019triton} GPU kernel for SSA. And we compare the forward and backward execution times of our SSA with those of FlashAttention-2~\cite{dao2023flashattention} across various number of tokens, using the implementation from Xformers~\cite{xFormers2022} for FlashAttention-2. The comparison results are shown in Figure~\ref{fig:speed}, which indicate that our SSA achieves comparable speeds to FlashAttention-2 when the number of tokens is low; however, as the number of tokens increases, the speed advantage of our SSA becomes more pronounced. Specifically, when the number of tokens reaches 128k, the forward and backward speeds of our SSA are $3.9\times$ and $9.6\times$ faster than those of FlashAttention-2, respectively, demonstrating the efficiency of our proposed SSA.

\vspace{12pt}
\noindent \textbf{Effectiveness of SSA.} We conduct ablation studies to validate the robustness of SSA. Given the insufficient geometric fidelity at $256^3$/ $384^3$ resolutions, which do not adequately reflect the model’s precision, and prohibitive computational costs at $1024^3$ resolution, we perform experiments at $512^3$ resolution. We establish three comparative configurations: 1) Full attention: directly training the DiT with full attention proves to be inefficient. Therefore, following Trellis' latent packing strategy~\cite{xiang2024structured}, we group latent tokens within $2^3$ local regions to reduce the number of tokens before feeding them into the DiT blocks. 2) NSA: process latent tokens as 1D sequences with fixed-length block partitioning, disregarding spatial coherence. 3) Our proposed SSA. The qualitative results are illustrated in Figure~\ref{fig:ablation_nsa}. It is evident that the full attention variant produces meshes with high-frequency surface artifacts, attributed to its forced packing operation that disrupts local geometric continuity. The NSA implementation exhibits training instability due to positional ambiguity in block partitioning, resulting in less smooth meshes. In contrast, our SSA not only preserves the details of the meshes, but also yields a smoother and more organized surface, thereby demonstrating the effectiveness of our proposed SSA mechanism.

\noindent\textbf{Effect of Sparse Conditioning Mechanism.}
We perform ablation experiments to validate the effect of the sparse conditioning mechanism at $512^3$ resolution. As demonstrated in Figure~\ref{fig:ablation_condition}, the exclusion of non-foreground conditioning tokens through sparse conditioning enables the generated meshes to achieve notably better alignment with the input images.

%% file: tables/dit_config.tex
\begin{table}
  \centering
  \caption{
        Training configurations for DiT at four voxel resolutions.  
        \textbf{Res.}, \textbf{NT}, \textbf{LR}, \textbf{BS}, and \textbf{TT} denote
        resolution, number of tokens, learning rate, batch size, and total training time, respectively.
    }
      \begin{tabular}{c  c c c c}
        \toprule
        \textbf{Res.} & \textbf{NT} & \textbf{LR} & \textbf{BS} & \textbf{TT} \\
        \midrule    
        $256^3$  & $\approx$2058 & 1e-4 & $8\times8$ & 2 days\\
        $384^3$  & $\approx$5510 & 1e-4 & $8\times8$ & 2 days\\
        $512^3$  & $\approx$10655 & 5e-5 & $8\times8$ & 2 days\\
        $1024^3$ & $\approx$45904 & 2e-5 & $2\times8$ & 1 day \\
        \bottomrule
      \end{tabular}
  \label{tab:dit_config}
\end{table}

%% file: tables/compare_dit.tex

\begin{table}[!t]
  \centering
  \caption{Quantitative comparisons of meshes generated by different methods in the image-to-3D task.}
  \resizebox{0.45\textwidth}{!}{%
  \begin{tabular}{l  c  c  c}
    \toprule
    \textbf{Methods} & \textbf{ULIP-2~$\uparrow$} & \textbf{Uni3D~$\uparrow$} & \textbf{OpenShape~$\uparrow$} \\
    \midrule    
    Trellis~\cite{xiang2024structured} & 0.2825 & 0.3755 & 0.1732 \\
    Hunyuan3D 2.0~\cite{zhao2025hunyuan3d} & 0.2535 & 0.3738 & 0.1699 \\
    TripoSG~\cite{li2025triposg} & 0.2626 & 0.3870 & 0.1728 \\
    Hi3DGen~\cite{ye2025hi3dgen} & 0.2725 & 0.3723 & 0.1689 \\
    \textbf{Ours} & \textbf{0.3111} & \textbf{0.3931} & \textbf{0.1752} \\
    \bottomrule
  \end{tabular}}
  \label{tab:comparison_i23d}
\end{table}

%% file: contents/5_conclusion.tex
\section{Conclusion}
\label{sec:conclusion}
\vspace{-6pt}
In this work, we presented a novel framework for high-resolution 3D shape generation, dubbed Direct3D-S2. The key contribution of our approach is the design of Spatial Sparse Attention (SSA) mechanism, which significantly accelerates the training and inference speed of DiT.
The integration of fully end-to-end symmetric sparse SDF VAE further enhances training stability and efficiency. Extensive experiments demonstrate that our Direct3D-S2 outperforms existing state-of-the-art image-to-3D methods in generation quality, while requiring only 8 GPUs for training.

\vspace{-4pt}
\section{Limitations}
\vspace{-6pt}
Our proposed spatial sparse attention achieves significant speed improvements over FlashAttention-2. However, the forward pass exhibits a notably smaller acceleration ratio compared to the backward pass. This discrepancy primarily stems from the computational overhead introduced by top-k sorting operations during the forward pass. We acknowledge this limitation and will prioritize optimizing these operations in future work.